\documentclass[sigconf, authorversion]{acmart}

\usepackage{subcaption}
\usepackage{svg}

\setcopyright{acmlicensed}
\copyrightyear{2025}
\acmYear{2025}
\setcopyright{acmlicensed}\acmConference[DigiPro '25]{The Digital Production
Symposium}{August 9, 2025}{Vancouver, BC, Canada}
\acmBooktitle{The Digital Production Symposium (DigiPro '25), August 9, 2025,
Vancouver, BC, Canada}
\acmDOI{10.1145/3744199.3744635}
\acmISBN{979-8-4007-2008-6/2025/08}
\acmSubmissionID{3580}

\begin{document}

\title{Automated Video Segmentation Machine Learning Pipeline}

\author{Johannes Merz}
\affiliation{%
  \institution{Image Engine Design Inc.}
  \city{Vancouver}
  \country{Canada}}
\email{johannesm@image-engine.com}

\author{Lucien Fostier}
\affiliation{%
  \institution{Image Engine Design Inc.}
  \city{Vancouver}
  \country{Canada}}
\email{lucienf@image-engine.com}




\begin{teaserfigure}
  \includegraphics[width=\textwidth]{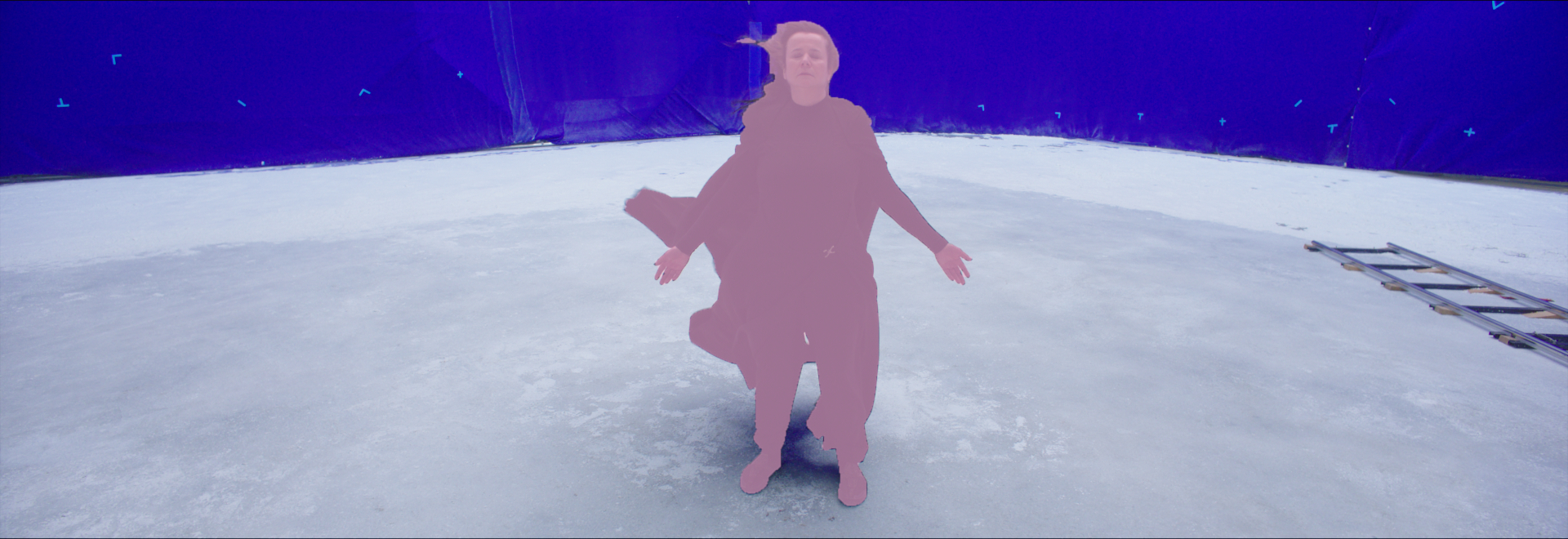}
  \caption{Mask output from a shot in \emph{Dune: Prophecy}, generated by our \emph{automatic video segmentation pipeline} and stored in our custom \emph{ObjectId} file format. \copyright \emph{HBO}}
  \label{fig:teaser}
\end{teaserfigure}

\begin{abstract}
Visual effects (VFX) production often struggles with slow, resource-intensive mask generation. This paper presents an automated video segmentation pipeline that creates temporally consistent instance masks. It employs machine learning for: (1) flexible object detection via text prompts, (2) refined per-frame image segmentation and (3) robust video tracking to ensure temporal stability. Deployed using containerization and leveraging a structured output format, the pipeline was quickly adopted by our artists. It significantly reduces manual effort, speeds up the creation of preliminary composites, and provides comprehensive segmentation data, thereby enhancing overall VFX production efficiency.
\end{abstract}

\maketitle

% Introduction
\section{Introduction}

Modern visual effects (VFX) production is characterized by tight turnaround times and increasingly animation-intensive workflows. Maintaining high quality under these pressures necessitates initiating animation work and delivering preview composites as early as possible in the schedule. 

These preview composites, which integrate computer graphic elements roughly composited onto background image plates, require masks to isolate foreground elements.

Generating these essential masks using existing traditional techniques, such as manual rotoscopy or bluescreen cleanup, proves notoriously time-consuming. Challenges like uneven bluescreens frequently necessitate significant manual effort to prepare masks. 

Importantly, much of this work is dedicated to producing assets that are ultimately disposable, serving only the immediate needs of temporary reviews or downstream department handovers. This results in a considerable waste of skilled resources on tasks that are only required for transient purpose, as these temporary masks often cannot be directly reused for final quality work due to differing requirements. 

These significant production challenges underscore a clear and pressing need for a more efficient, automated approach to mask generation in early-stage VFX workflows.

% Segmentation
\section{Automatic Segmentation Pipeline}
\label{sec:automaticPipeline}

Our main goal is to create a system that enables downstream departments to quickly pick up their respective tasks while minimizing the delay as much as possible. To achieve this, we developed a pipeline that can be run fully automatically without needing any user interaction. The default behaviour of this pipeline is designed to create an individual mask for each person in a plate, while being temporally consistent. The pipeline can be scheduled to run automatically when plates are ingested. Optionally, the default behaviour can be modified so that the pipeline generates masks for other objects and can then be run manually. 

After running the pipeline, the output is a temporally consistent instance segmentation. In general segmentation can be classified into semantic and instance segmentation. Semantic segmentation generates a mask that identifies all areas in an image that belong to the same class (e.g. people), but cannot differentiate between individual instances. This allows to e.g. replace backgrounds. Instance segmentation is a harder problem that can also differentiate between the individual instances in that image (e.g. identify individual people) and enables artists to place objects between characters.

\begin{figure*}
    \centering
    \includegraphics[width=1\linewidth]{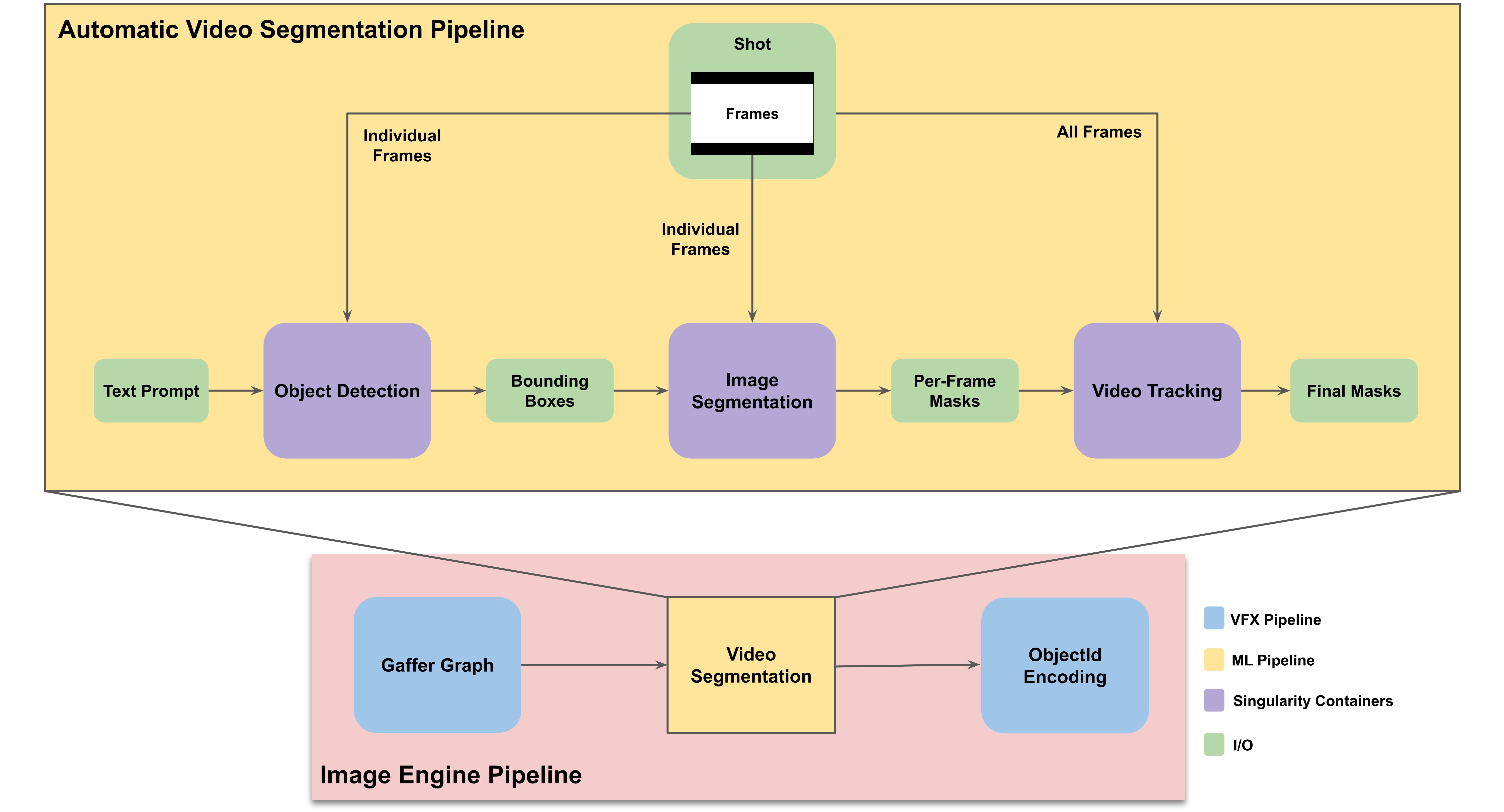}
     \caption{Overview of automatic video segmentation (see Section \ref{sec:automaticPipeline}) and integration into our VFX pipeline (see Section \ref{sec:integration}).}
    \label{fig:pipeline}
\end{figure*}

The pipeline consists of three fundamental stages as visualized in Figure \ref{fig:pipeline}:

\begin{enumerate}
	\item Object Detection
	\item Image Segmentation
	\item Video Tracking
\end{enumerate}

The following sections will describe each stage. Without limiting the generality we will assume that the pipeline is segmenting \emph{people} for the sake of simplicity of the description.

\subsection{Object Detection - Stage 1}

The first step in the pipeline performs object detection. This is the process of detecting all instances of a specific object in a shot. 

In machine learning (ML) multiple approaches for object detection exist. One type of neural network can be trained on specific categories (closed-set detection), which allows it to detect these categories with high accuracy. A popular system of this type is You Only Look Once (YoLo) \cite{redmon2016you}, which is pretrained on some categories, including \emph{“person”}. However, while testing YoLo on our shots, we quickly discovered that its limitations make it not a good fit for our use case in particular. Although it can detect people in general, the huge variety of possible human-like appearances in VFX can pose a challenge. For example, a character wearing a futuristic space helmet could potentially stump a pretrained base model due to the lack of similar samples in the public dataset the model has been trained on. Moreover, in VFX we mostly want to predict people, but not always. In order to modify the behaviour we would need to create a new dataset for each instance and thus significantly increase the manual work required.

To give the pipeline more flexibility without requiring data collection and fine-tuning for each category, we chose an open-set detection system instead, which generates a prediction given a conditioning input. Specifically, we chose to use GroundingDINO \cite{liu2024grounding}, which can be prompted using natural language. With the help of this object detection model, the artists can prompt the pipeline to generate masks for any object in the shot, simply by typing it out. Additionally, prompts can be concatenated which allows for the detection of different objects at the same time. For example a prompt could be \emph{"person, car"} to instruct the pipeline to generate individual masks for each person and each car across the whole shot. When our pipeline is run in the default automatic mode, the prompt \emph{“person”} is used. In this first step of the pipeline, the object detection is performed on each individual frame of the shot and outputs bounding boxes for each instance of the prompted objects.

\begin{figure*}
    \centering
    \includegraphics[width=1\linewidth]{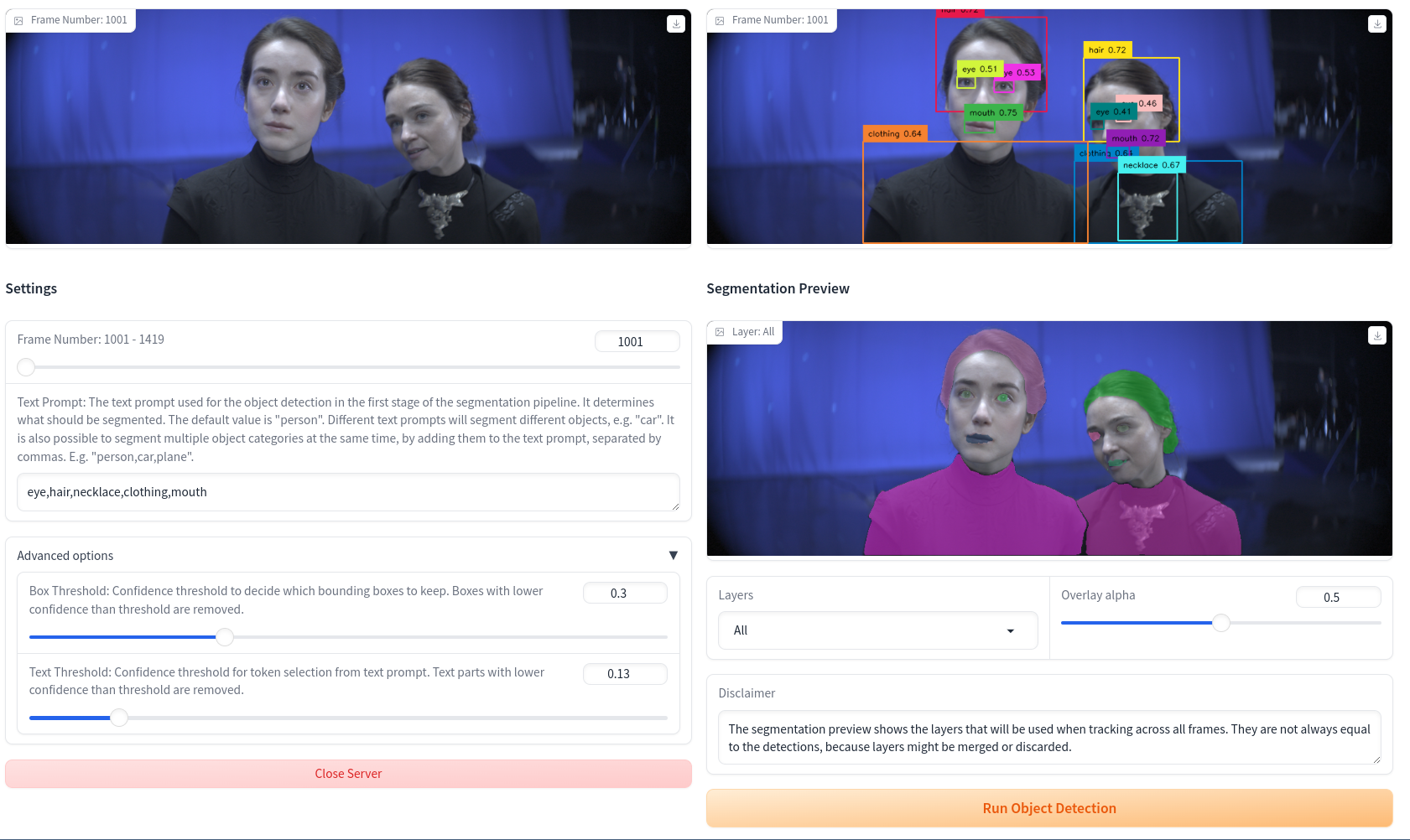}
     \caption{Object detection preview tool to quickly experiment with text prompts for stage 1 of the pipeline. It also shows a preview of the segmentation post-processing performed in stage 2. Images \copyright \emph{HBO}}
    \label{fig:odpreview}
\end{figure*}

The text prompt \emph{"person"} has proven to detect any human-like appearance with high probability and can therefore run unsupervised for the default case. Mask generation of other objects requires artists to manually edit the prompt. Since the pipeline tasks are dispatched to our render farm the interaction is not in real time, making it unfeasible to experiment with the text prompt by waiting for the final results. To solve this problem, we created a simple browser-based application to serve as a test bench for the artists (Figure \ref{fig:odpreview}) that provides an easy visualization of the bounding boxes. It also previews the results of stage 2, which is described in detail in Section \ref{sec:imageSegmentation}.

\subsection{Image Segmentation - Stage 2}
\label{sec:imageSegmentation}

\begin{figure}
     \centering
     \begin{subfigure}[b]{0.23\textwidth}
         \centering
         \includegraphics[width=\textwidth]{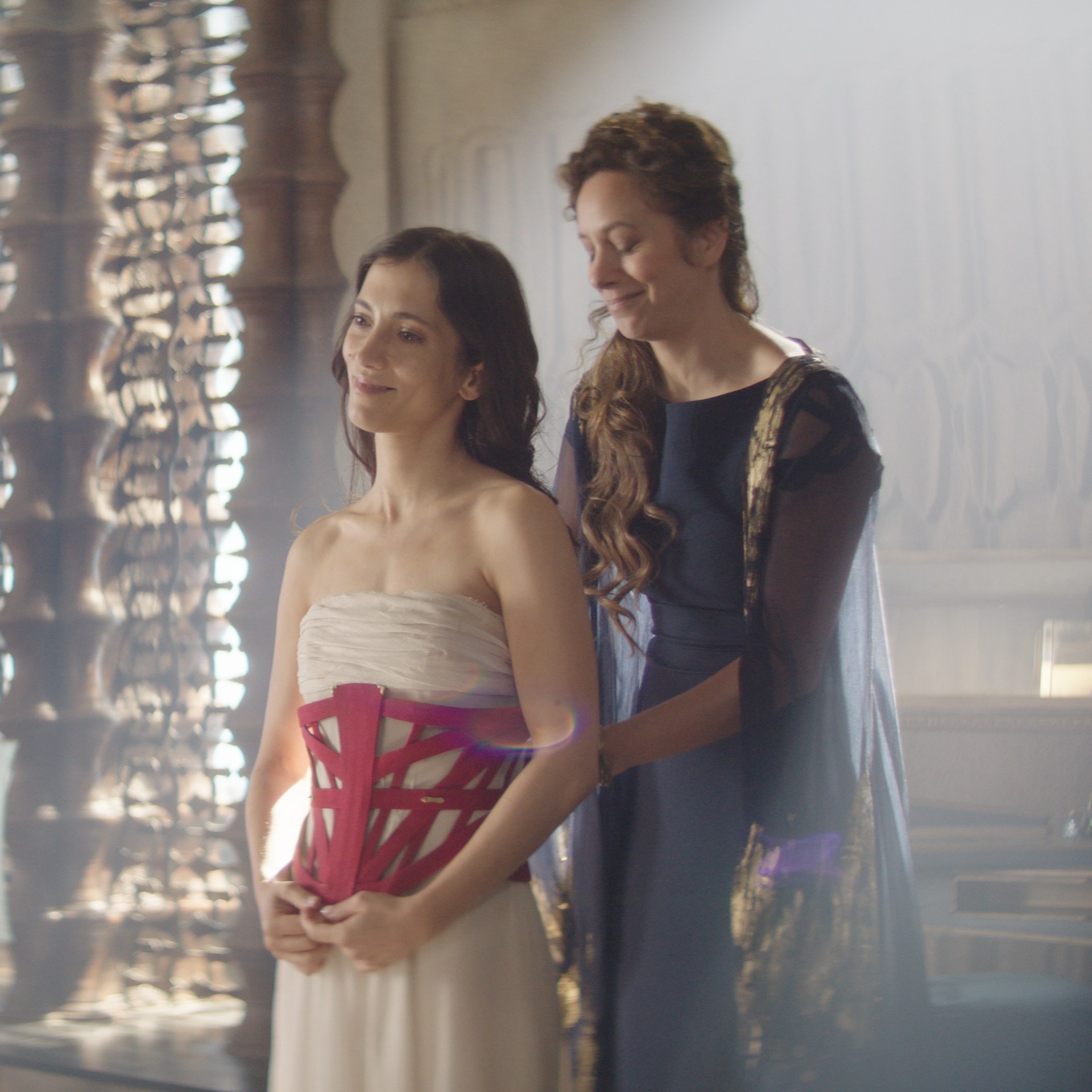}
         \caption{Plate}
         \label{fig:stage2_failure_a}
     \end{subfigure}
     \hfill
     \begin{subfigure}[b]{0.23\textwidth}
         \centering
         \includegraphics[width=\textwidth]{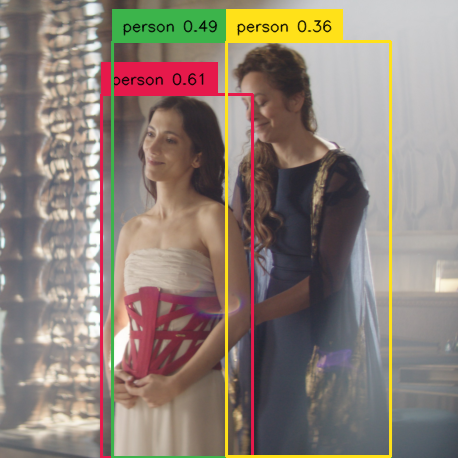}
         \caption{False-positive detection}
         \label{fig:stage2_failure_b}
    \end{subfigure}
    \begin{subfigure}[b]{0.23\textwidth}
         \centering
         \includegraphics[width=\textwidth]{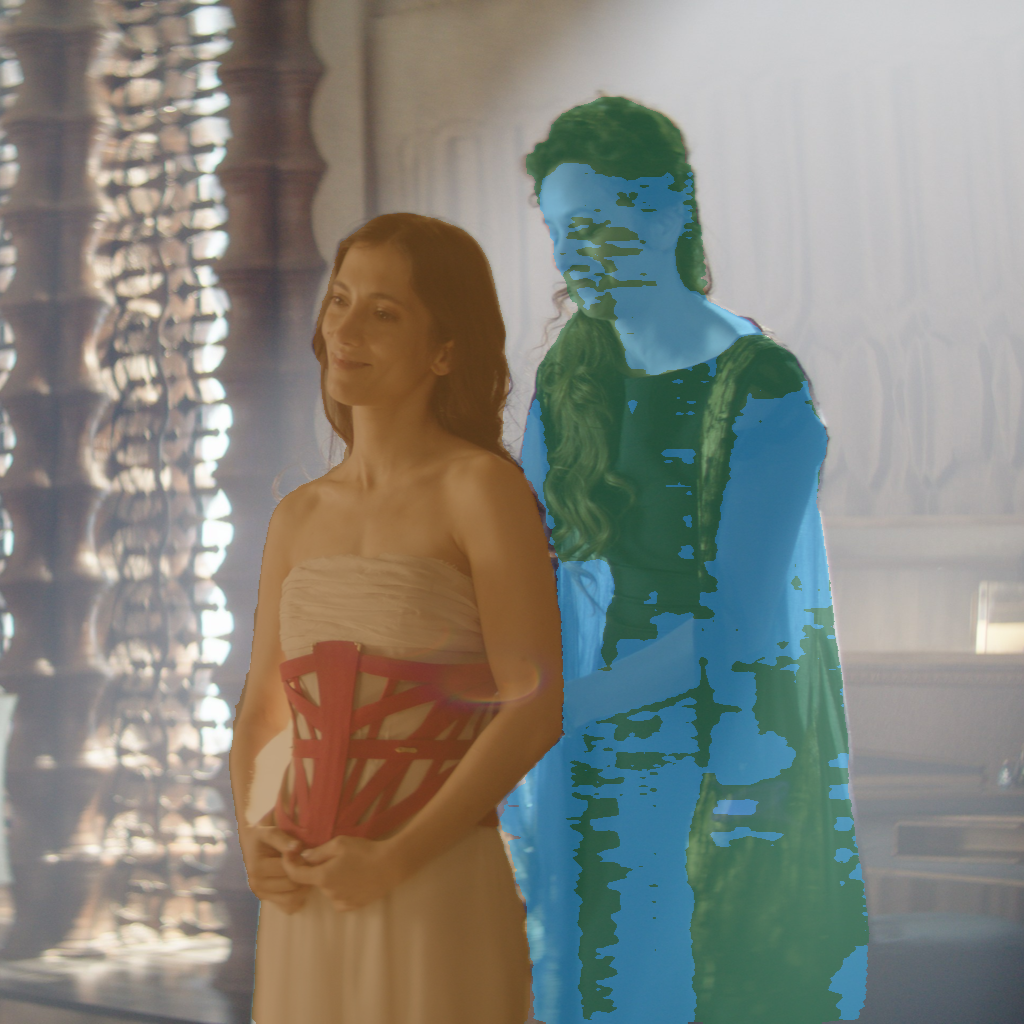}
         \caption{\emph{Without} post-processing}
         \label{fig:stage2_failure_c}
     \end{subfigure}
     \hfill
     \begin{subfigure}[b]{0.23\textwidth}
         \centering
         \includegraphics[width=\textwidth]{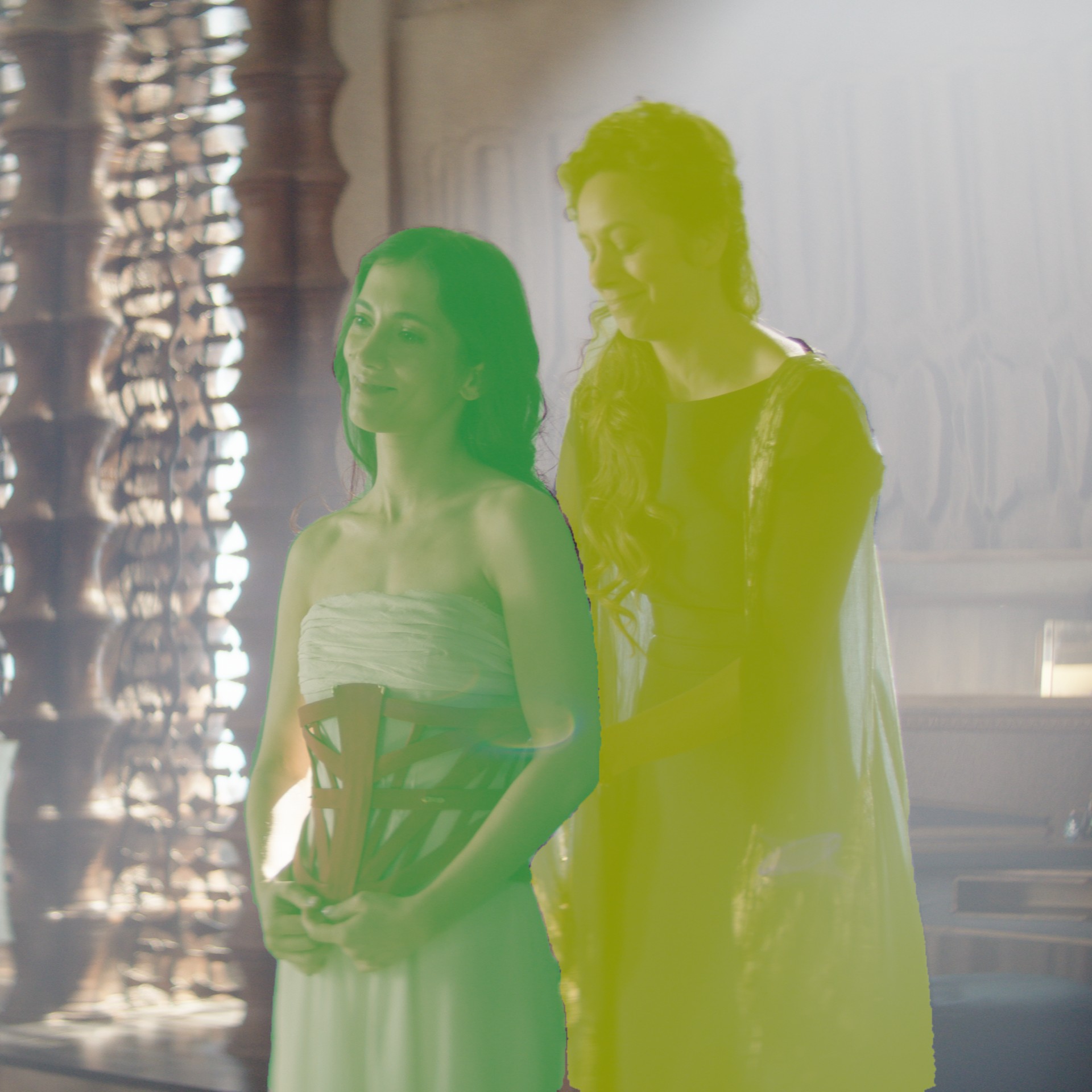}
         \caption{\emph{With} post-processing}
         \label{fig:stage2_failure_d}
     \end{subfigure}
     \caption{Example case where object detection (stage 1) detects a false-positive instance \textbf{(b)}. Directly passing these results to stage 3 leads to one person being split into two layers \textbf{(c)}. Using the post-processing in stage 2 resolves this problem by correctly discarding one layer \textbf{(d)}. Images \copyright \emph{HBO}}
     \label{fig:stage2_failure}
\end{figure}

The detections from the previous stage are a good first step for our pipeline, yet there are many potential issues that we need to address on a per-frame basis since the object detection is not guaranteed to be without fault:

\begin{enumerate}
	\item Sometimes the same person might be detected multiple times in a frame, or parts of the same person might be detected in separate channels.
	\item Especially in settings where people are closely intertwined (e.g. fight scenes), the object detection stage occasionally returns bounding boxes for the individual people, as well as a bounding box that combines both people or parts of both people, which is a false positive (Figure \ref{fig:stage2_failure}).
\end{enumerate}

Due to these challenges, we need to perform post-processing by merging or deleting some of the detection results, before we are able to properly address temporal consistency. Since the cases mentioned above are impossible to clean up based on the bounding boxes alone, we are performing the post-processing on segmentation mask results within these bounding boxes. 

To achieve this, we use a conditional segmentation model. These kinds of models predict a segmentation mask based on a conditioning input, in this case a bounding box. The most prominent models in this category are the Segment Anything Models (SAM), which can be prompted using bounding boxes and they predict a binary segmentation mask of the object enclosed by the bounding box. For initial experiments during the development of the pipeline we used the first iteration of SAM \cite{kirillov2023segment}, which we later replaced by the newly released SAM2 \cite{ravi2024sam} that also provides further benefits described in Section \ref{sec:stage3}.

Using SAM2, we first generate a segmentation mask for each detected bounding box on a given frame. Next, we calculate how much each mask overlaps with every other mask, based on the similarity metric 
\begin{equation}
    \text{Sim}(A, B) = \frac{A \cap B}{A},  \forall (A,B) \in Masks.
    \label{eq:asymDist}
\end{equation}
This metric answers the question: \textit{What proportion of A is covered by or intersects with B?} In other words, it measures the overlap with B from the perspective of A. Since the metric is asymmetric, which means that $\text{Sim}(A, B) \ne \text{Sim}(B, A)$, we are able to gain directional insights from it to answer questions such as: \textit{Which mask is part of another mask?} or \textit{Which mask is the result of a poor detection and groups unrelated parts of other people together?}

Using the calculated metrics we can discard and merge layers based on the following criteria:

\begin{enumerate}
	\item If a mask overlaps with at least two other masks with a threshold of $\text{Sim}(A, B) \ge 0.1$, it is very likely a false positive result from the object detection stage and can be discarded. The threshold to discard is low to guarantee that the layers that reach the next step are high-quality.
	\item If after deletion of the false positive layers there is still a very high overlap of a mask with exactly one other mask $\text{Sim}(A, B) \ge 0.6$, we merge the layers. The criterion to merge is much higher to ensure that we only merge layers that represent parts of the same person.
\end{enumerate}

\subsection{Video Tracking - Stage 3}
\label{sec:stage3}

The first two stages of the pipeline are working on a per-frame basis. That means that every result is calculated only based on the information available in one frame. Although the result of stage 2 is a set of plausible segmentation masks for each instance, there are still a few issues to address. 

Since each segmentation mask generated in stage 2 is based on the information of a single frame, there are slight differences between subsequent frames. These temporal inconsistencies lead to flickering across the whole shot. Because of the single frame context, the segmentation mask for the exact same person could also be in a different layer in frame $t$ than in frame $t+1$ and there is no direct way of relating the results of one frame to the results of subsequent frames.
Another issue is that the object detection might have missed people in a few frames that it has successfully detected in other frames, or that the segmentation mask predictions were too inaccurate and discarded in stage 2. This leads to missing frames for some layers.
Thus, at this point in the pipeline there is no temporal consistency yet.

For this task we use a different mode to operate SAM2, which is not only capable of generating masks for individual frames, but can also track prompts from one frame to subsequent frames in the same shot. Furthermore, in addition to being promptable with bounding boxes, it can also be prompted with masks directly. Thus, theoretically it would have been possible to prompt SAM2 directly with the bounding boxes from stage 1, however we found in our experiments that the per-frame post-processing performed in stage 2 is an essential factor to ensure the quality of the final results. Without this careful supervision, the SAM2 video tracking results quickly degrade if the bounding boxes are not placed ideally, which we cannot always rely on.

\subsubsection{Overall workflow}

The main idea of this stage is to prompt the video tracking model with the stage 2 segmentation masks from the first frame $I_{t=0}$ and propagate the layers forward for a fixed tracking interval with length $s_T$. The information received from this tracking step (which is the video prediction result by SAM2 $V_t$), is saved in a mask cache $M_t$  for each frame $t$ in the tracking interval by combining it with previous information that is already in the cache for this frame (Section \ref{sec:maskCache}). 

This process is repeated with a step size $s_P$ by comparing the information in the frame cache $M_{t+S_P}$ with the stage 2 segmentation masks at the same frame $I_{t+S_p}$ and using the combined information to re-prompt SAM2 (Section \ref{sec:promtpMask}). This is done to inject new high-quality information into the SAM2 prompt memory, because SAM2 tends to have diminished accuracy over longer tracking intervals, especially when multiple layers are tracked at the same time. Through experimentation we found that the values $s_P=5$ and $s_T=20$ provide an ideal balance between performance and quality.

During this merging process, we also determine whether the person in the stage 2 segmentation mask is already assigned to a specific layer in the mask cache, or if it needs to be added to a new layer. This allows us to keep the layers for each person consistent and add new people that only start appearing later in the shot.

Next, we repeat this process until the end of the shot is reached. Until now, each result SAM2 predicted only used prompt information from previous frames. However, due to the incremental step size $s_P$ it could be that new people were added a few frames after they actually appeared. Also sometimes the object detector in stage 1 does not immediately recognize people when they appear, because they might still be mostly occluded. To alleviate this, we now run the whole process in reverse direction. For this reverse process, we only need to use the stage 2 segmentation mask information of the very last frame, because all other necessary information has already been merged into the mask cache during the forward pass.

Finally, after running both the forward and the backward pass, the mask cache holds the final segmentation masks for each frame. These masks are both temporally smooth and individual people are in the same layer in across all frames.

\begin{figure}
    \centering
    \includegraphics[width=1\linewidth]{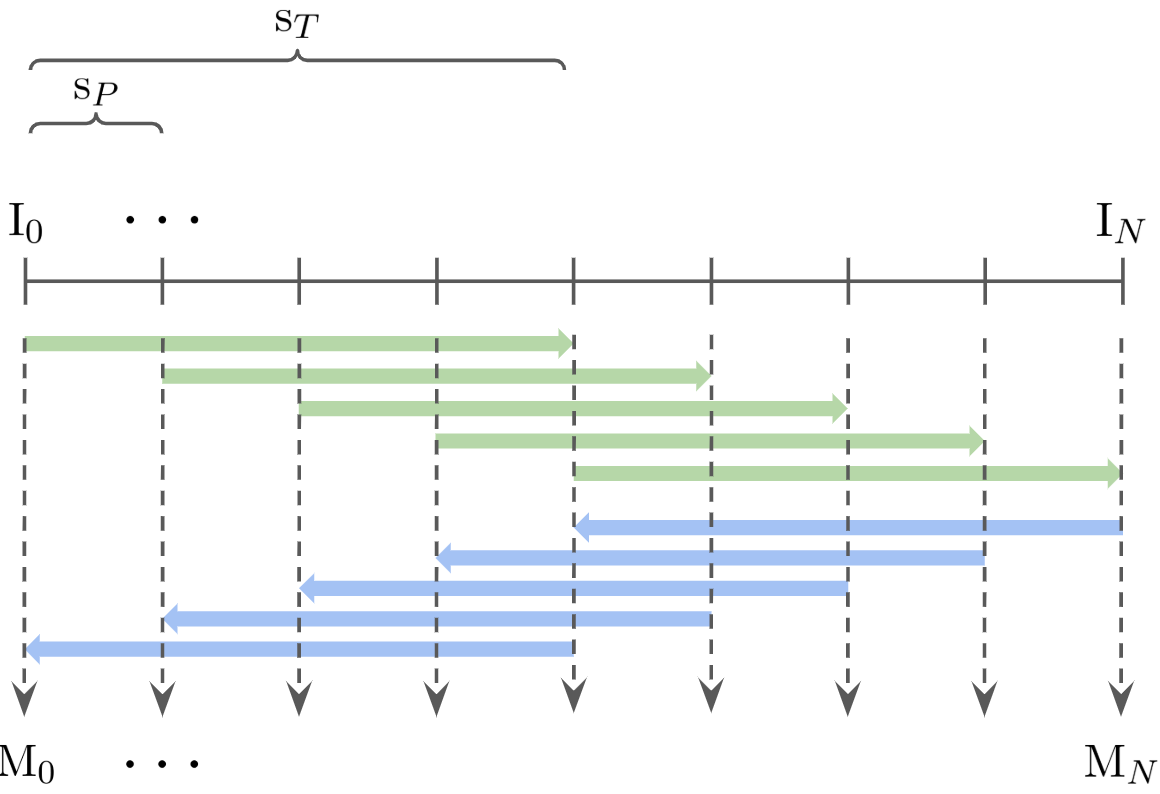}
     \caption{Visualization of the video tracking workflow. Stage 2 segmentations $I$ are tracked and results of all iterative tracking steps are merged into the mask cache $M$.}
    \label{fig:automaticWorkflow}
\end{figure}

\begin{figure*}
     \centering
     \begin{subfigure}[b]{\textwidth}
         \centering
         \includegraphics[width=\textwidth]{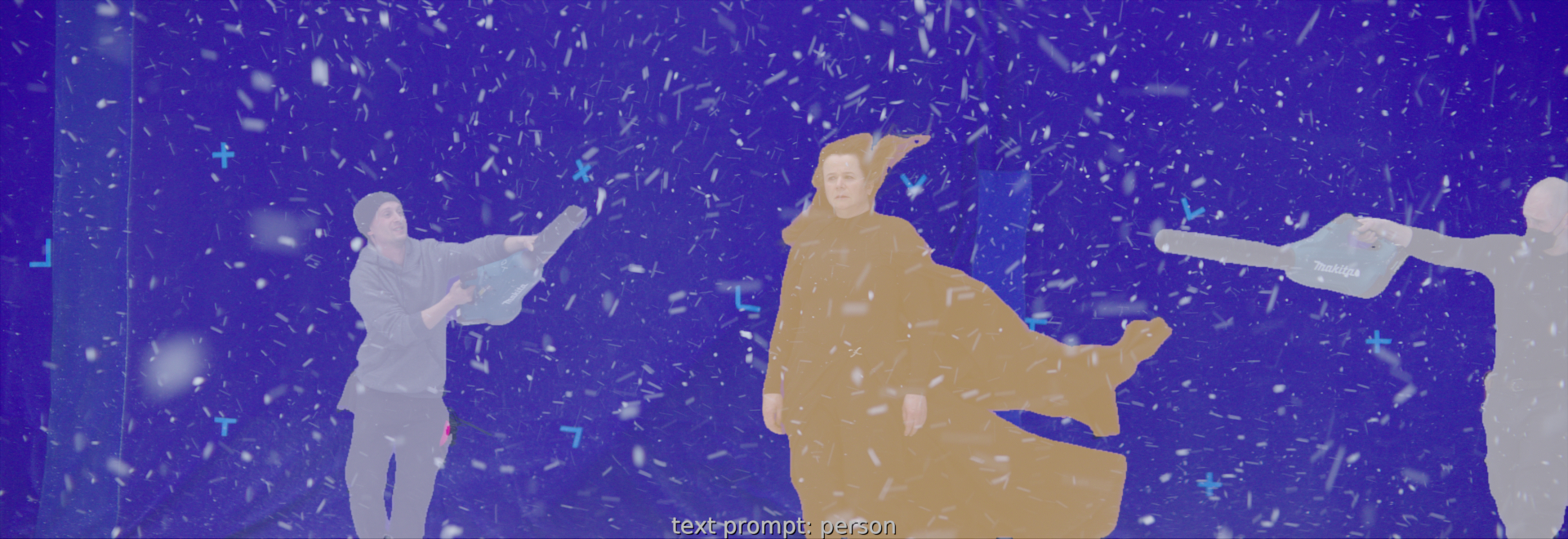}
         \caption{Text prompt \emph{"person"}.}
    \end{subfigure}
    \begin{subfigure}[b]{0.49\textwidth}
         \centering
         \includegraphics[width=\textwidth]{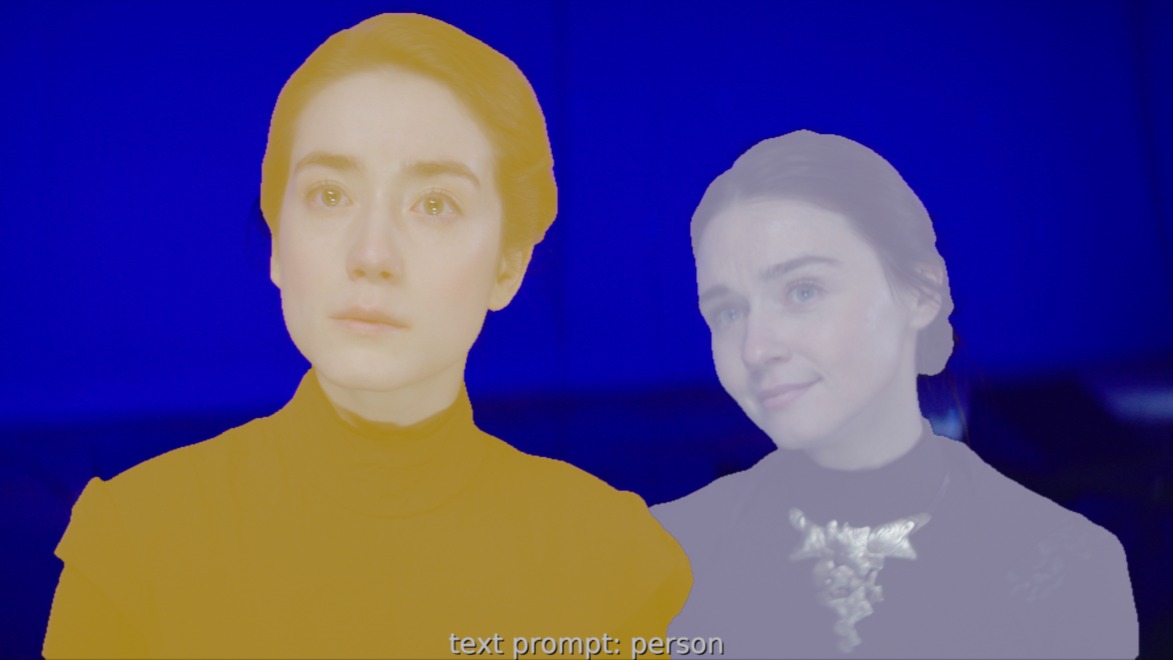}
         \caption{Text prompt \emph{"person"}.}
     \end{subfigure}
     \hfill
     \begin{subfigure}[b]{0.49\textwidth}
         \centering
         \includegraphics[width=\textwidth]{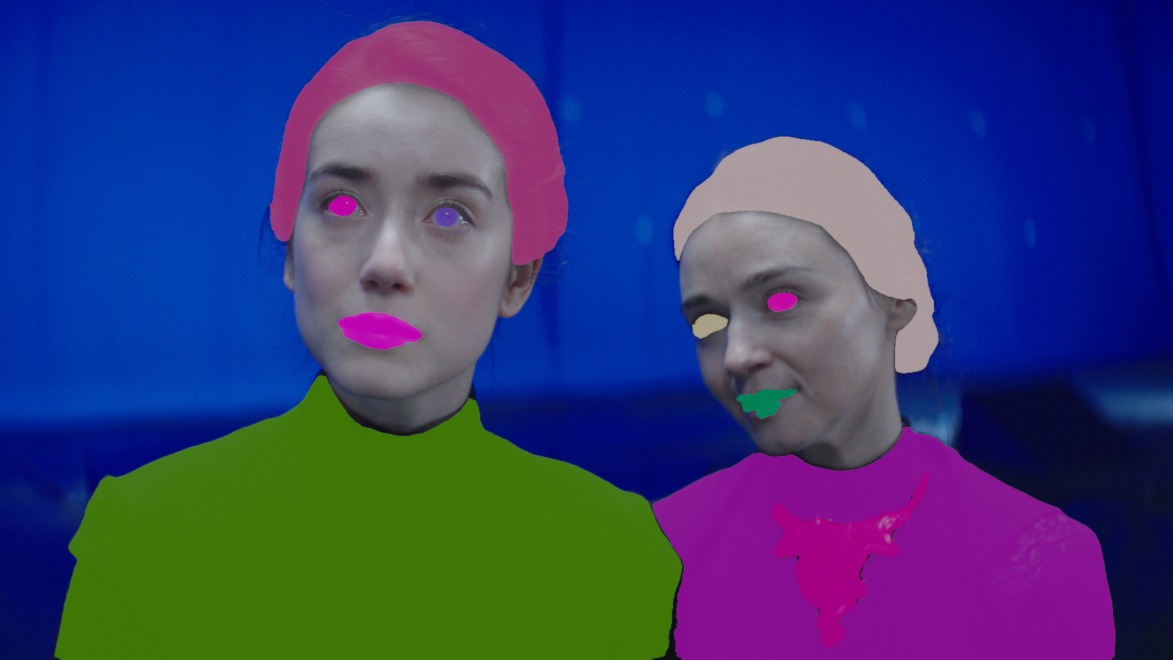}
         \caption{Text prompt \emph{"eye, hair, necklace, clothing, mouth"}.}
     \end{subfigure}
     \caption{Example results created with the automatic pipeline different text prompts. The complete results can be found in the supplementary video. \copyright \emph{HBO}}
     \label{fig:results}
\end{figure*}

Figures \ref{fig:teaser} and \ref{fig:results} and the supplementary video show examples of final results created by the pipeline. 

\subsubsection{Combining Information in the Mask Cache}
\label{sec:maskCache}

As mentioned, whenever SAM2 returns predicted segmentation masks, they are merged with any previously existing information for this frame in the mask cache. The cache holds two 2D-matrices for each frame $t$ with the same dimensions as the input frames of the shot. 

The first matrix $M_{ID, t}$ holds the layer ID that each pixel is assigned to. For example, if there are two people detected in the frame, then each pixel corresponding to the first person is set to 1, each pixel corresponding to the second person is set to 2, and each pixel that corresponds to background is set to 0. The second matrix $M_{P, t}$ holds the probability that the pixel belongs to the layer ID in $M_{ID, t}$ for each pixel, which is the prediction that SAM2 returns.

In case the mask cache for this frame is empty, we store the exact result from SAM2. Otherwise, we compare the result of SAM2 with the information in the mask cache. For each pixel, we decide if SAM2 provides a new prediction with a higher probability than currently assumed. To prioritize information that is already in the cache, we require the new prediction from SAM2 to be higher by a margin of $\epsilon = 0.1$ before it is allowed to overwrite information in the cache. This means information will only be updated for pixels where $M_{P, t} + \epsilon \leq \text{V}_{P, t}$. In these cases the model is very confident that it has new information for this frame that is more accurate than the current state.

\subsubsection{Generating New Prompt Mask}
\label{sec:promtpMask}

At each step $s_P$ we generate a new prompt mask for SAM2 by merging the mask cache information with the stage 2 segmentation masks. In the best case scenario, both sets of masks are highly similar. We check this case first by using the metric Intersection over Union (IoU) defined by 
\begin{equation}
    \text{IoU}(M, I) = \frac{M \cap I}{M \cup I}.
\end{equation}
This metric tells us how similar two masks are in relation to their combined size. If this number is 1, they are identical. If it is 0, there is no overlap. For each layer in the stage 2 masks, we check if it has an $\text{IoU} \ge 0.9$ with any layer already in the mask cache, resulting in a direct match.

In case there is no direct match, we investigate further with a similar approach to the mask post-processing in stage 2, using the asymmetric metric defined in Equation \eqref{eq:asymDist}. The asymmetry allows us to determine which masks are enclosing another mask and are a partial match. This happens for example, if the mask cache only has a part (arm) but the stage 2 information at the next step is more complete (whole person). Just like before, if the stage 2 segmentation mask overlaps with more than one layer in the mask cache, we discard it.

Finally, we use the content of the mask cache as the template for the new SAM2 prompt and merge the layers with direct and partial matches of the stage 2 segmentation masks. This ensures that each person has a consistent layer across the shot. Note that this prompt mask is only used to guide the next tracking step, it is not saved in the mask cache. Only SAM2 tracking results are saved in the mask cache to ensure that all the final information is coming from the same model, giving us smooth temporal consistency.

\section{Interactive Segmentation}
\label{sec:interactiveSegmentation}

The previously described pipeline is designed to be as hands-off and automatic as possible. The artists can only interact with the pipeline by changing the text prompt. Given that the object detection in stage 1 results in bounding boxes, not every object can be perfectly represented. For example, if an object is very large and in the background of a shot, the bounding box might be very large and enclose a number of other objects that the artist did not intend to add. In these cases it becomes very hard for the image segmentation model to predict which object is exactly defined by the bounding box. Small details can also create difficult cases.

In practice, the automatic pipeline is sufficient for most cases. However, we leveraged another input modality of the SAM2 model to provide the artists with an alternative approach for these edge cases. This mode puts the artist in direct control of the segmentation masks that are then being tracked.

Section \ref{sec:stage3} describes that SAM2 can be prompted using bounding boxes and segmentation masks. Alternatively, it accepts positive and negative points as inputs. These points are simply 2D pixel positions that tell the model if the respective point is supposed to be inside of the segmentation mask for this frame (positive point) or outside (negative point). The model then predicts a segmentation mask with the help of this information.

Based on this possibility, we created a separate browser-based application that allows the artists to define these points by selecting them directly on the frame. The system then predicts the segmentation mask for this frame and the artist can iteratively fine-tune the results.

\begin{figure*}
    \centering
    \includegraphics[width=1\linewidth]{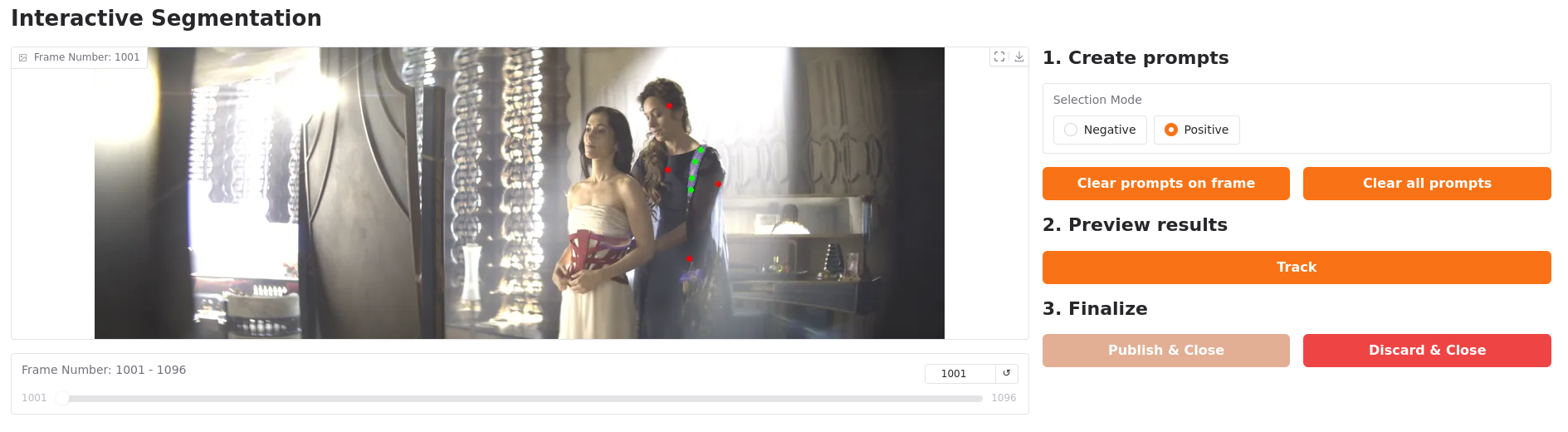}
     \caption{The interactive segmentation workflow can be used to quickly generate masks in collaboration with the SAM2 model. Image \copyright \emph{HBO}}
    \label{fig:interactiveSegmentation}
\end{figure*}

Figure \ref{fig:interactiveSegmentation} shows an example where a few positive (green) and negative (red) clicks are sufficient to create a mask for a fine detail on the dress of the women. 

Once the artist has generated all the masks, they are then used as prompt masks for the video tracking in our pipeline. This interactive tool is only one of the ways artists can start the video tracking task with manual mask prompts. They can also use any other traditional VFX tool to create a mask on one or multiple keyframes of a shot and submit it to our video tracking system to propagate the information through the shot.

% Integration
\section{Pipeline Integration}
\label{sec:integration}

The development of our automatic video segmentation pipeline involved integrating machine learning libraries within existing traditional VFX toolsets as shown in Figure \ref{fig:pipeline}.

Initially, our approach used \emph{subprocess} calls from within our VFX Digital Content Creation (DCC) software to execute scripts leveraging ML libraries, necessitating the use of a different Python interpreter than the DCC's native one. 

While this allowed for the development of early versions of this pipeline, we quickly discovered that this method was time-consuming to maintain; any updates to the pipeline or the underlying ML models required substantial changes to the required ML libraries within the specific script environments. 

\subsection{Singularity Integration}

To address this dependency and maintenance challenge, we transitioned to using container technology to package the ML libraries, our custom automatic segmentation pipeline scripts and the models together. This containerized approach enabled our ML engineers to autonomously craft container images containing all the necessary libraries and tools specific to a particular model, enabling rapid deployment within our VFX pipeline.

We evaluated several containerization options, including Docker \cite{DockerPage}, but found that its requirement for \emph{superuser} permissions was a limitation, as our ML processes on the render farm operate under the user profiles that initiated the jobs. 

Singularity/Apptainer \cite{ApptainerPage} proved to be a more flexible solution for our needs and offered excellent support for hardware acceleration with Nvidia GPUs. 

The adoption of Singularity has been instrumental in enabling the rapid and streamlined deployment of model updates throughout our VFX production pipeline.

\subsection{ObjectId}

In 2012 during the post-production on \emph{World War Z} at Cinesite \cite{CinesitePage}, ObjectId was developed to be able to color grade crowds of zombies per character instance.
At the time of development, there were no open source or render integrated solutions and Cryptomatte \cite{CryptomattePage} was not presented until SIGGRAPH 2015. 

Thanks to Cinesite's and Image Engine’s partnership since 2015, ObjectId has been integrated into Image Engine's Arnold rendering pipeline and since 2021 into our automatic segmentation pipeline.

ObjectId’s format is designed to support an arbitrary number of samples per pixel, which allows representing complex scenarios like overlapping objects or semi-transparent regions. 

The format also stores a manifest of all the masks per frame stored within the file metadata. This manifest serves to map internally used numeric identifiers to user-friendly object identifiers, typically representing hierarchy locations or descriptive names, thus providing a clear link between the raw pixel data and the scene structure.

Each pixel sample stores information about its coverage, represented as an alpha value and a numeric ID. This numeric ID is the link that connects to the user-friendly name or hierarchy location defined in the manifest, allowing it to be matched using a user-defined filter.

This design enables users to define a specific filter based on object identifiers, which is then matched against the identifier associated with every pixel sample. A final, usable mask for the filtered object(s) is then produced by accumulating the opacity (alpha) of all the pixel samples that match the user's defined filter.

Using this structured mask storage format is highly beneficial as it allows us to maintain the information linking the user's initial prompting (identifying the class or specific object) all the way through to the final output of the pipeline.

\begin{figure*}[ht]
    \centering
    \includegraphics[width=1\linewidth]{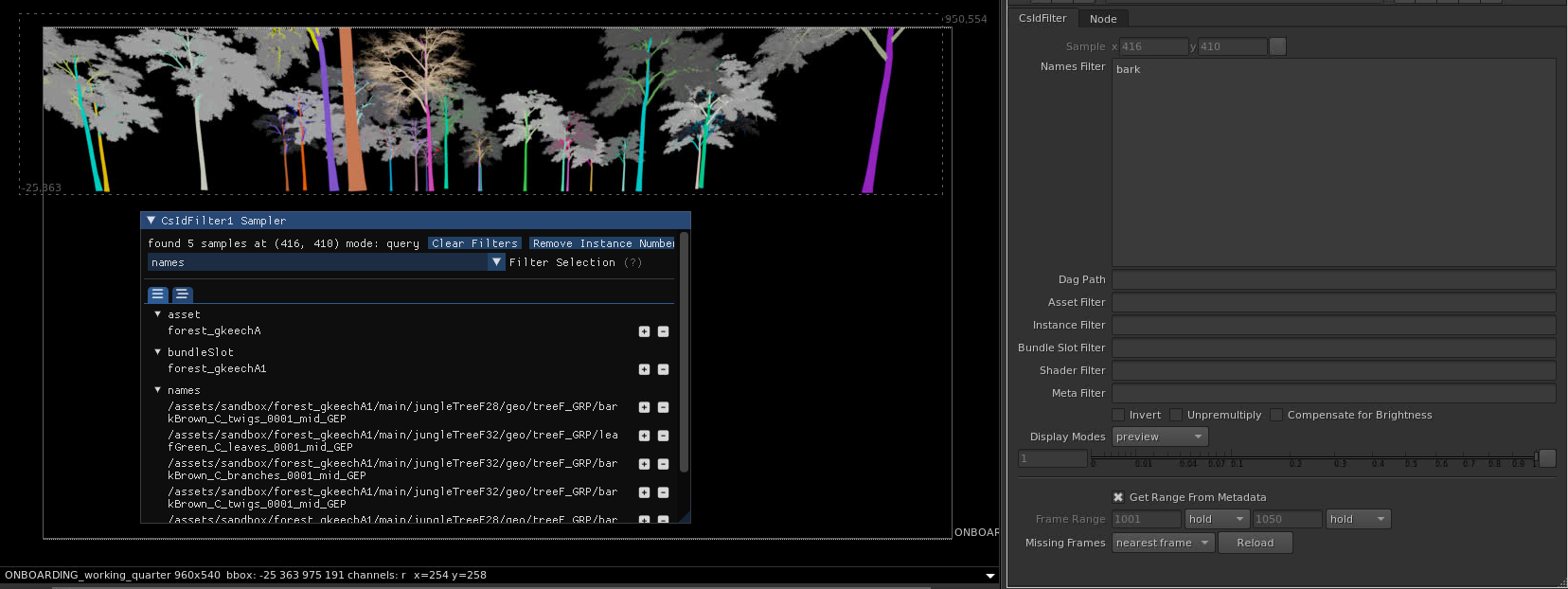}
    \caption{Mask picking using ObjectId custom filtering node within \copyright Foundry's Nuke. \copyright \emph{Cinesite}}
    \label{fig:objectIdNuke}
\end{figure*}

The inherent structured nature of the output and its consistency across different shots significantly enables template work, streamlining downstream processes. This approach provides a rich, layered representation of the scene's segmentation that can be flexibly queried to extract specific object masks and supports more automated, predictable workflows (Figure \ref{fig:objectIdNuke}).

% Production Impact
\section{Production Impact}
\label{sec:productionImpact}

% \subsection{Existing Workflow}

The integration of our automatic video segmentation pipeline significantly impacted our production workflows, primarily by streamlining the creation of WIP (Work-in-Progress) composites and \emph{precomps}.

Existing methods for generating masks for these temporary outputs are notoriously time-consuming. 
Temporary masks produced by rotoscoping often cannot be reused directly for final quality work, because the required breakdown of shapes and details differs significantly \cite{InvisibleArtOfRoto}, requiring artists to start from scratch rather than refining existing splines for high-quality results. 

Therefore, much of this effort is dedicated to producing assets that are ultimately disposable and serve only the immediate needs of temporary reviews or downstream department handovers. This results in a considerable waste of skilled resources on tasks that are only required for transient purposes.

\subsection{Unique Benefits of the Automatic Segmentation}

Our automatic video segmentation pipeline offers a distinct advantage over traditional methods by being inherently more efficient and requiring fewer dedicated resources to achieve comprehensive results.

We have found that our solution provides greater flexibility in obtaining complete masks for elements within a plate, a significant improvement compared to traditional workflows where resource limitations often dictate that effort is concentrated only on the absolutely necessary areas of an image.

This constraint with traditional approaches frequently led to under-requesting masks that could have been valuable elsewhere in the pipeline if they had been readily available.

The core advantage of our new approach is its ability to deliver more extensive coverage of what can be extracted from the plate, coupled with a greater diversity in the types of masks we can generate.

For instance, our system allows us to easily obtain detailed sub-part masks for complex objects (see Figure \ref{fig:odpreview}), whereas the traditional, resource-bound approach often only focuses on a silhouette mask within the same time and budget constraints.

This enhanced capability provides downstream departments with a richer set of segmentation data, potentially unlocking new creative possibilities and further enhancing the overall impact on VFX workflow
scheduling.
Beyond the inefficiencies on these types of tasks, the traditional approach to video segmentation significantly impacts production scheduling and resource allocation. 

VFX production schedules often necessitate that work begins concurrently across multiple departments relatively early in the process, particularly in \emph{Layout}, \emph{Animation}, \emph{Fx} and \emph{Lighting}. A primary objective during these early phases is the ability to quickly assemble and present sequences or key shots to the client for review and feedback. Achieving these early client presentations is critical for locking down creative direction and ensuring the project is on the right track.

Furthermore, committing valuable rotoscoping resources early in a show's life cycle can severely reduce flexibility for the rest of production, making it challenging to adapt to evolving creative demands or unforeseen complexities. 

Manual rotoscoping scheduling can bring challenges by either being over scheduled early on when requests are made broadly across many potential elements, leading to wasteful allocation of resources, or under scheduled, resulting in significant scheduling bottlenecks and challenges during the critical final stages when a high volume of precise segmentation requests often converges on the roto/paint team.

Recognizing this, some supervisors strategically prefer to delay requesting compositing and roto resources until absolutely necessary to avoid this early-stage waste, underscoring the need for more efficient solutions that can provide quick, temporary masks without heavily impacting the schedule or budget early on. 

Our automatic segmentation pipeline achieves this goal with an average throughput of 10 frames per minute across its three stages, with the most computationally intensive operations being per-frame segmentation and tracking. This efficiency allows for rapid iteration, enabling the generation of a new version of a typical 100-frames VFX shot within just 10 minutes. See Table \ref{tab:processingTimes} for more details.

\begin{table}
\caption{Randomly selected samples of processing times for each step of the pipeline in minutes.}
\begin{tabular}{|c|c|c|c|c|}
\hline
\textbf{Frames} & \textbf{Stage 1} & \textbf{Stage 2} & \textbf{Stage 3} & \textbf{Total}\\
\hline
283 & 1:44 & 11:58 & 9:44 & 23:26\\
205 & 1:28 & 11:57 & 8:50 & 22:05\\
173 & 1:13 & 5:58  & 4:29 & 11:40\\
139 & 1:05 & 4:56  & 3:54 & 9:55 \\
\hline
\end{tabular}
\label{tab:processingTimes}
\end{table}

\subsection{Rapid Adoption}

\begin{figure}
    \centering
    \begin{subfigure}[b]{0.47\textwidth}
         \centering
         \includegraphics[width=\textwidth]{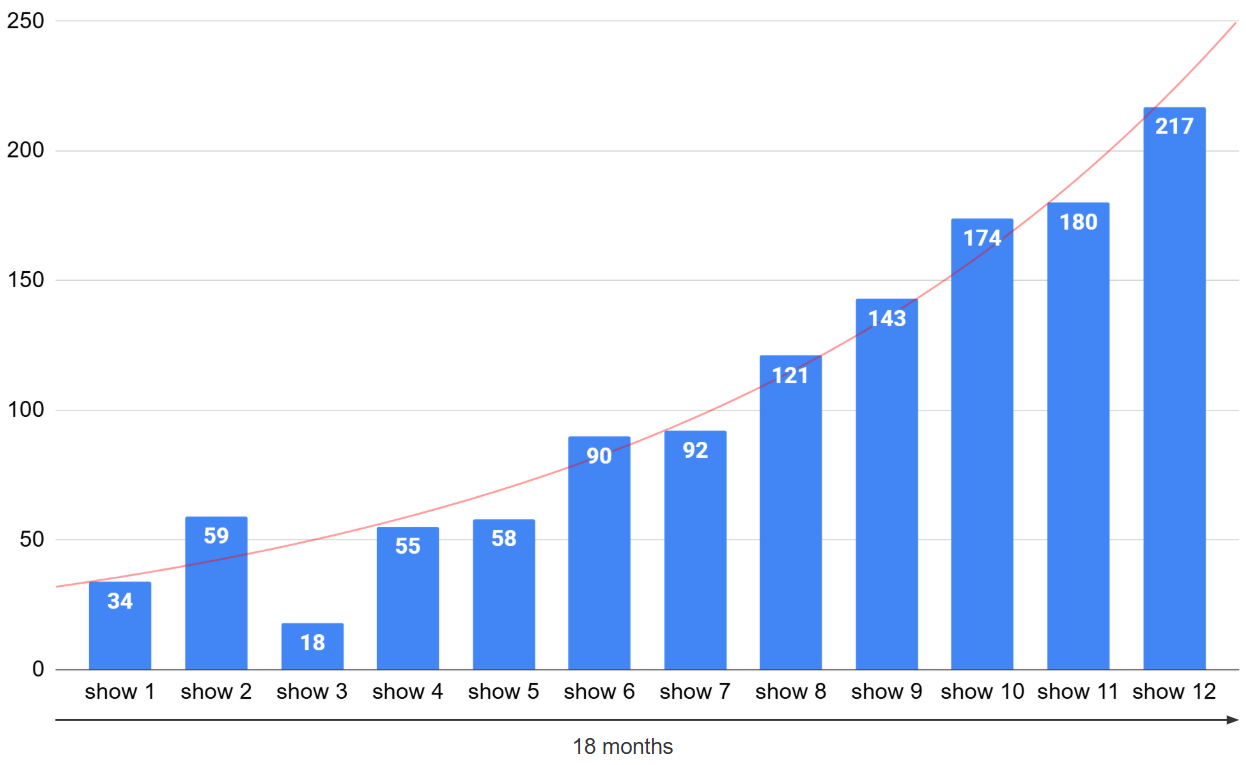}
         \caption{Amount of shots using our automatic segmentation per show (2024-2025)}
         \label{fig:showsAdoption}
    \end{subfigure}
    \begin{subfigure}[b]{0.47\textwidth}
         \centering
         \includegraphics[width=\textwidth]{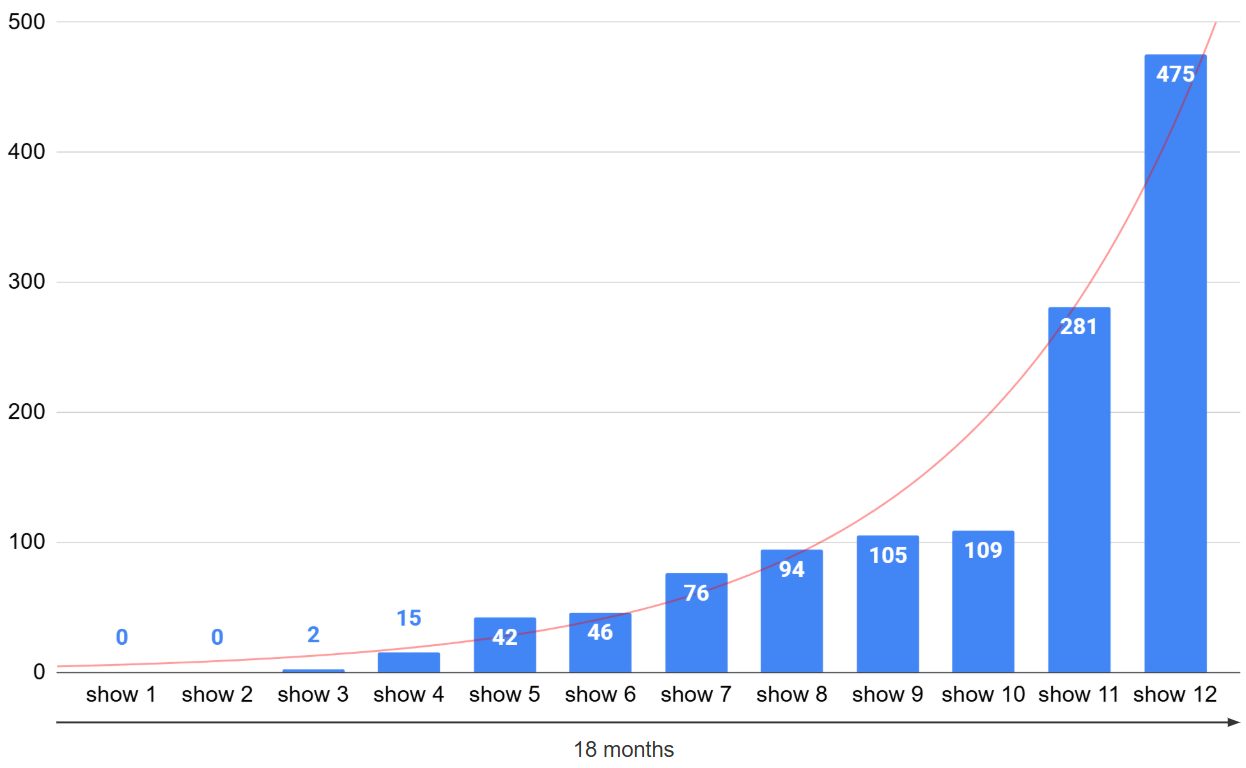}
         \caption{Production image sequences using automatic segmentation as a source (2024-2025)}
         \label{fig:Downstream department adoption}
    \end{subfigure}
    \caption{Statistics of the automatic segmentation pipeline}
\end{figure}

Since its integration into our production pipeline, the automatic video segmentation system has been successfully deployed on 12 different shows, processing a total of 1241 shots to date (Figure \ref{fig:showsAdoption}).

We have observed a significant level of adoption of the automatically generated segmentation masks by downstream departments, clearly evidenced by the substantial number of image sequences produced by these teams that use our masks as a source for compositing and other tasks (Figure \ref{fig:Downstream department adoption}).

Feedback from artists has been particularly encouraging; members of our compositing team report requiring only minimal adjustments, typically tweaking the automatically generated video segmentation in just 10\% to 15\% of cases. 

This low percentage of required manual intervention demonstrates the solidity of the results, as shown in the simplicity of the post-processing required in their Nuke scenes used to create a pre-multiplied version of the plate (see the supplementary material for details). This further highlights that the output is of sufficient quality to be used directly or with only minor modifications.

% Limitations
\section{Limitations}

After releasing the segmentation pipeline it has quickly established itself as an important tool at Image Engine. However, it does not come without limitations.

\begin{figure*}[!ht]
    \centering
    \includegraphics[width=1\linewidth]{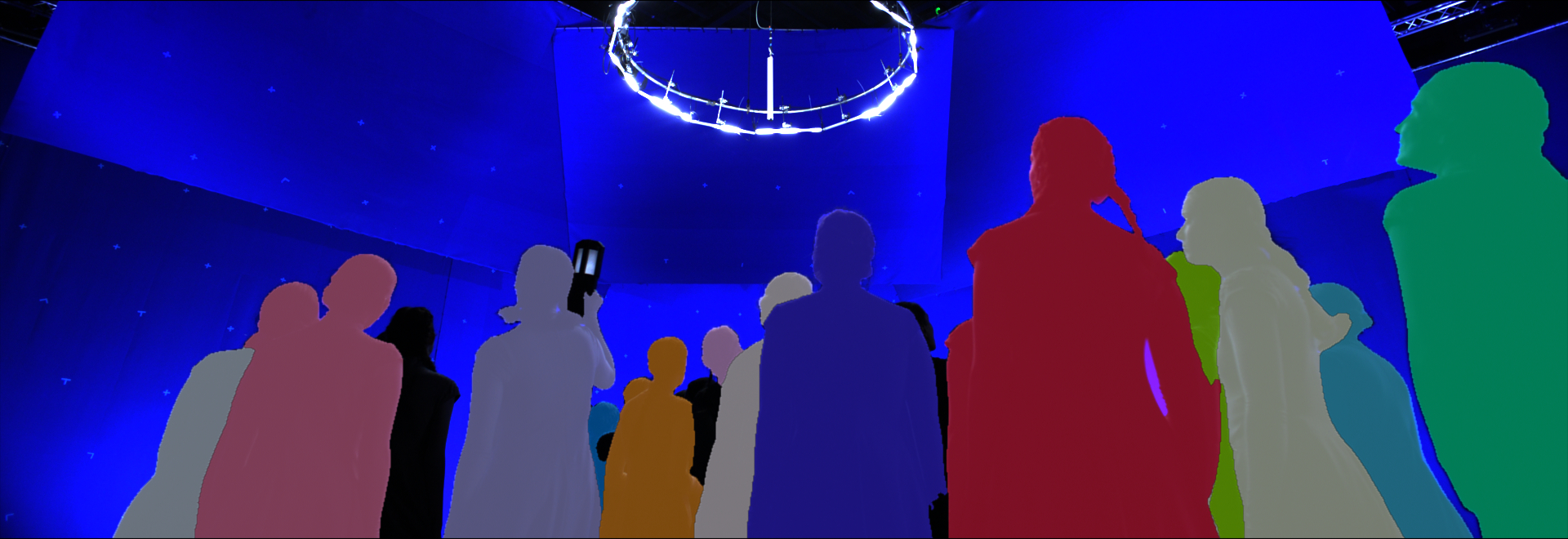}
    \caption{A result of the automatic pipeline, where some layers were dropped to prevent interference during the tracking stage. \copyright \emph{HBO}}
    \label{fig:limitation}
\end{figure*}

During the post-processing in stage 2 (Section \ref{sec:imageSegmentation}) some layers might be dropped that might have been acceptable on closer manual inspection. If these specific layers are needed, the artists would need to rerun the automatic pipeline with a more targeted text prompt, or use the interactive segmentation approach (Section \ref{sec:interactiveSegmentation}). Figure \ref{fig:limitation} shows a result that produced many layers with good quality for different people. However, some people in the far background were not segmented in any layer. These layers would have interacted too much with the rest of the information and might have degraded the tracking results. This happens, because in general the post-processing prioritizes the quality and stability of the remaining layers over completeness to ensure a high usability for downstream tasks.

Since the pipeline is based on machine learning models, using the pipeline comes with a significant GPU hardware requirement. Each execution is scheduled on dedicated GPU servers equipped with NVIDIA A4000 GPUs with 16GB of GPU memory. To minimize GPU memory usage, each stage of the pipeline is executed individually. This allows each model to use the full memory. However, even then the video tracking with SAM2 can quickly reach the memory limit. Due to this limitation, we currently limit the video tracking by resizing the shots so that the maximum dimension is scaled to 1024 pixels while keeping the aspect ratio intact.
% Future work
\section{Conclusion \& Future Work}
\label{sec:futureWork}

This paper presents a fully automatic segmentation pipeline designed to address the demanding needs of VFX production. Its successful implementation had a demonstrated high impact at our studio and led to practical benefits and efficiency gains in our artists' workflows. This work represents a significant step in our goal to use machine learning in a very selective manner to address bottlenecks in our pipeline.

The successful integration opens several avenues for future exploration, including tighter integration into our VFX environment and extending the ability of the pipeline to also track splines.
 
\subsection{GafferML}

Building upon our initial pipeline integration, more recent developments aim to enable deeper integration of ML workflows directly within our DCC environments. 

In particular, we are actively developing GafferML, a project which wraps the ONNX \cite{ONNX} framework into dedicated nodes within Gaffer's \cite{GafferDigipro} various processing contexts. 

While this work is currently in its early development, we have already achieved promising results, successfully integrating and experimenting with some of our ML models directly inside Gaffer to perform image processing tasks such as image segmentation, depth estimation, and object detection. 

Beyond image-based tasks, we have also conducted experiments with geometry processing, specifically utilizing the SMPL model \cite{SMPL:2015} to generate posed mesh of human bodies. 

Looking ahead, our future work includes plans to integrate our interactive segmentation and tracking capabilities directly into Gaffer nodes to enhance human-in-the-loop systems that require artist interaction and refinement.

\subsection{Spline Tracking}

While our initial pipeline integration is primarily focused on servicing departments requiring disposable video segmentation for client presentation material, a key area for future work is dedicated to improving the efficiency of the workflow for high-quality segmentation originating from our rotoscopy department. 

For this effort to be successful, we recognize the need to pivot from solutions focused solely on per-pixel segmentation of input images towards predicting and tracking rotoscopy splines, as this constitutes the main data format utilized by this department. 

Therefore, our aim is to develop specific machine learning models designed to leverage information extracted from both input image and temporal data to accurately predict spline knot locations and curve characteristics.

This approach is intended to significantly save time and resources during the critical phase of executing high-quality rotoscopy work for final shot completion.

\bibliography{main.bib}

%%% -*-BibTeX-*-
%%% Do NOT edit. File created by BibTeX with style
%%% ACM-Reference-Format-Journals [18-Jan-2012].

\begin{thebibliography}{12}

%%% ====================================================================
%%% NOTE TO THE USER: you can override these defaults by providing
%%% customized versions of any of these macros before the \bibliography
%%% command.  Each of them MUST provide its own final punctuation,
%%% except for \shownote{}, \showDOI{}, and \showURL{}.  The latter two
%%% do not use final punctuation, in order to avoid confusing it with
%%% the Web address.
%%%
%%% To suppress output of a particular field, define its macro to expand
%%% to an empty string, or better, \unskip, like this:
%%%
%%% \newcommand{\showDOI}[1]{\unskip}   % LaTeX syntax
%%%
%%% \def \showDOI #1{\unskip}           % plain TeX syntax
%%%
%%% ====================================================================

\ifx \showCODEN    \undefined \def \showCODEN     #1{\unskip}     \fi
\ifx \showDOI      \undefined \def \showDOI       #1{#1}\fi
\ifx \showISBNx    \undefined \def \showISBNx     #1{\unskip}     \fi
\ifx \showISBNxiii \undefined \def \showISBNxiii  #1{\unskip}     \fi
\ifx \showISSN     \undefined \def \showISSN      #1{\unskip}     \fi
\ifx \showLCCN     \undefined \def \showLCCN      #1{\unskip}     \fi
\ifx \shownote     \undefined \def \shownote      #1{#1}          \fi
\ifx \showarticletitle \undefined \def \showarticletitle #1{#1}   \fi
\ifx \showURL      \undefined \def \showURL       {\relax}        \fi
% The following commands are used for tagged output and should be
% invisible to TeX
\providecommand\bibfield[2]{#2}
\providecommand\bibinfo[2]{#2}
\providecommand\natexlab[1]{#1}
\providecommand\showeprint[2][]{arXiv:#2}

\bibitem[Apptainer(2025)]%
        {ApptainerPage}
\bibfield{author}{\bibinfo{person}{Apptainer}.} \bibinfo{year}{2025}\natexlab{}.
\newblock
\newblock
\urldef\tempurl%
\url{https://apptainer.org/}
\showURL{%
\tempurl}


\bibitem[BorisFX(2021)]%
        {InvisibleArtOfRoto}
\bibfield{author}{\bibinfo{person}{BorisFX}.} \bibinfo{year}{2021}\natexlab{}.
\newblock \bibinfo{booktitle}{\emph{The invisible art of roto}}.
\newblock
\urldef\tempurl%
\url{https://blog.borisfx.com/the-invisible-art-of-roto}
\showURL{%
\tempurl}


\bibitem[Cinesite(2025)]%
        {CinesitePage}
\bibfield{author}{\bibinfo{person}{Cinesite}.} \bibinfo{year}{2025}\natexlab{}.
\newblock
\newblock
\urldef\tempurl%
\url{https://www.cinesite.com}
\showURL{%
\tempurl}


\bibitem[Docker(2025)]%
        {DockerPage}
\bibfield{author}{\bibinfo{person}{Docker}.} \bibinfo{year}{2025}\natexlab{}.
\newblock
\newblock
\urldef\tempurl%
\url{https://www.docker.com/}
\showURL{%
\tempurl}


\bibitem[Foundation(2025)]%
        {ONNX}
\bibfield{author}{\bibinfo{person}{The~Linux Foundation}.} \bibinfo{year}{2025}\natexlab{}.
\newblock \bibinfo{title}{Open Neural Network Exchange (ONNX)}.
\newblock
\newblock
\urldef\tempurl%
\url{https://onnx.ai/}
\showURL{%
\tempurl}


\bibitem[Friedman and Jones(2015)]%
        {CryptomattePage}
\bibfield{author}{\bibinfo{person}{Jonah Friedman} {and} \bibinfo{person}{Andrew~C. Jones}.} \bibinfo{year}{2015}\natexlab{}.
\newblock \bibinfo{title}{Fully automatic ID mattes with support for motion blur and transparency.}
\newblock
\newblock
\urldef\tempurl%
\url{https://github.com/Psyop/Cryptomatte}
\showURL{%
\tempurl}


\bibitem[Haddon et~al\mbox{.}(2016)]%
        {GafferDigipro}
\bibfield{author}{\bibinfo{person}{John Haddon}, \bibinfo{person}{Andrew Kaufman}, \bibinfo{person}{David Minor}, \bibinfo{person}{Daniel Dresser}, \bibinfo{person}{Ivan Imanishi}, {and} \bibinfo{person}{Paulo Nogueira}.} \bibinfo{year}{2016}\natexlab{}.
\newblock \showarticletitle{Gaffer: an open-source application framework for VFX}. In \bibinfo{booktitle}{\emph{Proceedings of the 2016 Symposium on Digital Production (DigiPro)}}. \bibinfo{pages}{31--42}.
\newblock


\bibitem[Kirillov et~al\mbox{.}(2023)]%
        {kirillov2023segment}
\bibfield{author}{\bibinfo{person}{Alexander Kirillov}, \bibinfo{person}{Eric Mintun}, \bibinfo{person}{Nikhila Ravi}, \bibinfo{person}{Hanzi Mao}, \bibinfo{person}{Chloe Rolland}, \bibinfo{person}{Laura Gustafson}, \bibinfo{person}{Tete Xiao}, \bibinfo{person}{Spencer Whitehead}, \bibinfo{person}{Alexander~C Berg}, \bibinfo{person}{Wan-Yen Lo}, {et~al\mbox{.}}} \bibinfo{year}{2023}\natexlab{}.
\newblock \showarticletitle{Segment anything}. In \bibinfo{booktitle}{\emph{Proceedings of the IEEE/CVF International Conference on Computer Vision (ICCV)}}. \bibinfo{pages}{4015--4026}.
\newblock


\bibitem[Liu et~al\mbox{.}(2024)]%
        {liu2024grounding}
\bibfield{author}{\bibinfo{person}{Shilong Liu}, \bibinfo{person}{Zhaoyang Zeng}, \bibinfo{person}{Tianhe Ren}, \bibinfo{person}{Feng Li}, \bibinfo{person}{Hao Zhang}, \bibinfo{person}{Jie Yang}, \bibinfo{person}{Qing Jiang}, \bibinfo{person}{Chunyuan Li}, \bibinfo{person}{Jianwei Yang}, \bibinfo{person}{Hang Su}, {et~al\mbox{.}}} \bibinfo{year}{2024}\natexlab{}.
\newblock \showarticletitle{Grounding dino: Marrying dino with grounded pre-training for open-set object detection}. In \bibinfo{booktitle}{\emph{European Conference on Computer Vision (ECCV)}}. Springer, \bibinfo{pages}{38--55}.
\newblock


\bibitem[Loper et~al\mbox{.}(2015)]%
        {SMPL:2015}
\bibfield{author}{\bibinfo{person}{Matthew Loper}, \bibinfo{person}{Naureen Mahmood}, \bibinfo{person}{Javier Romero}, \bibinfo{person}{Gerard Pons-Moll}, {and} \bibinfo{person}{Michael~J. Black}.} \bibinfo{year}{2015}\natexlab{}.
\newblock \showarticletitle{{SMPL}: A Skinned Multi-Person Linear Model}.
\newblock \bibinfo{journal}{\emph{ACM Trans. Graphics (Proc. SIGGRAPH Asia)}} \bibinfo{volume}{34}, \bibinfo{number}{6} (\bibinfo{date}{Oct.} \bibinfo{year}{2015}), \bibinfo{pages}{248:1--248:16}.
\newblock


\bibitem[Ravi et~al\mbox{.}(2024)]%
        {ravi2024sam}
\bibfield{author}{\bibinfo{person}{Nikhila Ravi}, \bibinfo{person}{Valentin Gabeur}, \bibinfo{person}{Yuan-Ting Hu}, \bibinfo{person}{Ronghang Hu}, \bibinfo{person}{Chaitanya Ryali}, \bibinfo{person}{Tengyu Ma}, \bibinfo{person}{Haitham Khedr}, \bibinfo{person}{Roman R{\"a}dle}, \bibinfo{person}{Chloe Rolland}, \bibinfo{person}{Laura Gustafson}, {et~al\mbox{.}}} \bibinfo{year}{2024}\natexlab{}.
\newblock \showarticletitle{Sam 2: Segment anything in images and videos}.
\newblock \bibinfo{journal}{\emph{arXiv preprint arXiv:2408.00714}} (\bibinfo{year}{2024}).
\newblock


\bibitem[Redmon et~al\mbox{.}(2016)]%
        {redmon2016you}
\bibfield{author}{\bibinfo{person}{Joseph Redmon}, \bibinfo{person}{Santosh Divvala}, \bibinfo{person}{Ross Girshick}, {and} \bibinfo{person}{Ali Farhadi}.} \bibinfo{year}{2016}\natexlab{}.
\newblock \showarticletitle{You only look once: Unified, real-time object detection}. In \bibinfo{booktitle}{\emph{Proceedings of the IEEE Conference on Computer Vision and Pattern Recognition (CVPR)}}. \bibinfo{pages}{779--788}.
\newblock


\end{thebibliography}

\end{document}